\documentclass[conference]{IEEEtran}
\IEEEoverridecommandlockouts
\usepackage{cite}
\usepackage{amsmath,amssymb,amsfonts}
\usepackage{algorithm}
\usepackage{algorithmic}
\usepackage{graphicx}
\usepackage{textcomp}
\usepackage{xcolor}
\usepackage{float}
\usepackage{balance}

\def\BibTeX{{\rm B\kern-.05em{\sc i\kern-.025em b}\kern-.08em
    T\kern-.1667em\lower.7ex\hbox{E}\kern-.125emX}}
\begin{document}

\makeatletter
\newcommand\fs@norules{\def\@fs@cfont{\bfseries}\let\@fs@capt\floatc@ruled
  \def\@fs@pre{}%
  \def\@fs@post{}%
  \def\@fs@mid{\kern3pt}%
  \let\@fs@iftopcapt\iftrue}
\makeatother
\floatstyle{norules}
\restylefloat{algorithm}

\title{SafeSwarm: Decentralized Safe RL for the Swarm of Drones Landing in Dense Crowds\\
\thanks{Research reported in this publication was financially supported by the RSF grant No. 24-41-02039.}
}


\author{\IEEEauthorblockN{Grik Tadevosyan}
\IEEEauthorblockA{\textit{Skoltech}\\
Moscow, Russia \\
grik.tadevosyan@skoltech.ru}
\and
\IEEEauthorblockN{Maksim Osipenko}
\IEEEauthorblockA{\textit{Skoltech}\\
Moscow, Russia \\
maksim.osipenko@skoltech.ru}
\and
\IEEEauthorblockN{Demetros Aschu}
\IEEEauthorblockA{\textit{Skoltech}\\
Moscow, Russia \\
demetros.tareke@skoltech.ru}
\and
\IEEEauthorblockN{Aleksey Fedoseev}
\IEEEauthorblockA{\textit{Skoltech}\\
Moscow, Russia \\
aleksey.fedoseev@skoltech.ru}
\and
\IEEEauthorblockN{Valerii Serpiva}
\IEEEauthorblockA{\textit{Skoltech}\\
Moscow, Russia \\
valerii.serpiva@skoltech.ru}
\and
\IEEEauthorblockN{Oleg Sautenkov}
\IEEEauthorblockA{\textit{Skoltech}\\
Moscow, Russia \\
oleg.sautenkov@skoltech.ru}
\and
\IEEEauthorblockN{Sausar Karaf}
\IEEEauthorblockA{\textit{Skoltech}\\
Moscow, Russia \\
sausar.karaf@skoltech.ru}
\and
\IEEEauthorblockN{Dzmitry Tsetserukou}
\IEEEauthorblockA{\textit{Skoltech}\\
Moscow, Russia \\
d.tsetserukou@skoltech.ru}
}


\maketitle

\begin{abstract}

This paper introduces a safe swarm of drones capable of performing landings in crowded environments robustly by relying on Reinforcement Learning techniques combined with Safe Learning. The developed system allows us to teach the swarm of drones with different dynamics to land on moving landing pads in an environment while avoiding collisions with obstacles and between agents. 

The safe barrier net algorithm was developed and evaluated using a swarm of Crazyflie 2.1 micro quadrotors, which were tested indoors with the Vicon motion capture system to ensure precise localization and control.

Experimental results show that our system achieves landing accuracy of  2.25 cm with a mean time of 17 s and collision-free landings, underscoring its effectiveness and robustness in real-world scenarios. This work offers a promising foundation for applications in environments where safety and precision are paramount.

\end{abstract}


\section{Introduction}
In recent years, Safe Reinforcement Learning (Safe RL) has emerged as a critical methodology in robotics, driving advances in systems that require both adaptability and robustness, especially in uncertain and dynamic environments. Safe RL represents a significant departure from traditional Reinforcement Learning (RL) approaches, which lack intrinsic mechanisms to ensure safety during exploration \cite{xiao2024}, \cite{xiao2023}. Such limitations highlight the need for improved frameworks in applications where operational safety and adaptability are paramount.

Although alternative approaches, e.g., Model Predictive Control (MPC) combined with Safe Learning, offer certain safety guarantees, they often fall short in dynamic and complex environments. MPC-based methods can struggle with slow adaptation times, which is problematic when the environment is replete with unknown and moving obstacles, such as pedestrians, where accurate position data are not available. In scenarios with a high number of states or complex environments, the performance of MPC can substantially degrade, leading to safety risks \cite{astghik2023_1}, \cite{koo2015}, \cite{astghik2023second}.

\begin{figure}[t]
\centering
\includegraphics[width=0.9\linewidth]{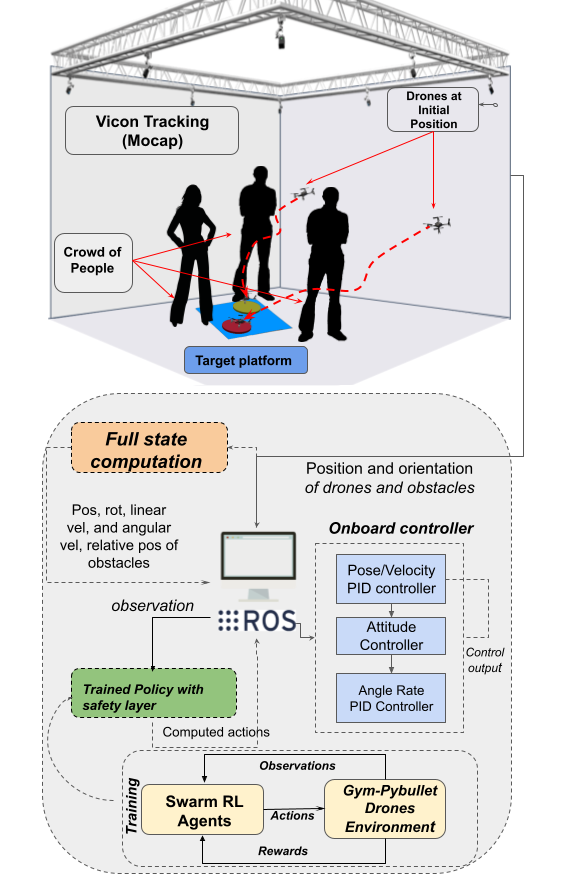}
\caption{The SafeSwarm system overview.}
\label{fig:title}
\end{figure}

One problem for MPC is the number of agents, for MPC each agent should know dynamics of other agents and environment, however, in real problems agents do not know the dynamics of other agents and environment \cite{lapandic2021}. \\
Furthermore, as the complexity of multi-agent systems grows, many algorithms exhibit scalability issues, making them unsuitable for swarm robotics where multiple agents must operate simultaneously and coordinate effectively \cite{lowe2017}. To address these challenges, we propose an approach that integrates an adaptive and robust learning algorithm designed for multi-agent environments with obstacles. Our solution leverages the Multi-Agent Proximal Policy Optimization (MAPPO) algorithm, equipped with a Multi-Layer Perceptron (MLP) neural network architecture tailored to the needs of decentralized multi-agent systems. 
Another method of avoiding obstacles is to use differential optimization-based CBF \cite{dai2023}, however, this method works for the convex and strongly convex cases.


\section{Related Works}

In recent years, research on autonomous drones and swarm robotics has surged, with a strong emphasis on enhancing the robustness and safety of these systems. In addition, articles on drone landing with Reinforcement Learning increase 4 times \cite{Amendola2024}. Very interesting work on single-drone landing on a moving platform in a disturbed environment is \cite{peter2024}, where landing pads make complex motions, and \cite{ladosz2024}, which is done using vision-based deep-reinforcement learning. Another work, which has been done to land for a swarm of drones, is \cite{aschu2024}. However, none of these works has solved the problem for dynamic or cluttered environments. An effective approach to planning drone swarm paths in cluttered environments is presented in \cite{zhao2024}, however, the spatial distribution of the vehicle is determined by the trajectory planner itself. In \cite{falanga2017} a vision-based landing concept is proposed where only the camera onboard was applied for high-speed drone positioning; however, the accuracy of such a vision-based model may not be sufficient for multi-agent systems.

As multiagent systems for drones continue to evolve, there is an increasing need for agents capable of navigating environments while avoiding obstacles and coordinating with each other to achieve safe landings. This increased focus on safety is particularly crucial in scenarios where swarms of drones operate nearby, as agents must not only avoid obstacles but also prevent collisions with each other during complex landing maneuvers.
Transferring these models from a simulation to the physical world \cite{Loquercio_2020} introduces challenges, including limitations in the accuracy of a simulation when reflecting real-world dynamics, noise, and unaccounted variables. These discrepancies can impact the robustness of UAV systems when faced with environmental disturbances, such as wind or sensor noise, that are difficult to accurately replicate in simulations. Model-free solutions often face scalability issues, especially for large-scale multiagent systems \cite{Shukla_2022}. Thus, increasing the number of agents in a swarm can lead to computational overhead and communication bottlenecks, impacting real-time decision making and control. 

Several works, including \cite{lin2024} and \cite{wang2024}, address obstacle avoidance through RL frameworks. However, these models can struggle in highly dynamic environments with moving obstacles or unpredictable changes. The reliance on DRL models that are trained on specific scenarios limits their generalization to new unseen conditions. Meanwhile, real-world applications often require adaptability to various dynamic and complex scenarios. The MorphoLander system applies the single-agent PPO to achieve a decentralized multi-agent landing \cite{sausar2023}. Although it demonstrates high precision, the approach did not take into account internal and external collisions between the swarm agents. The MARLander \cite{aschu2024} framework demonstrates effective local path planning within controlled simulations. However, its ability to generalize to highly dynamic real-world environments with unpredictable obstacles or rapidly changing conditions is limited. The framework primarily trains in static environments, which may not fully replicate the complexity of real-world scenarios. As a result, the model's adaptability to environments with dynamic obstacles, such as moving objects or varying environmental factors, may be limited. 

Safety mechanisms, for example in \cite{yuan2022}, could provide constraints to minimize collision risks. However, they are often heuristic-based and may not guarantee strict compliance under all conditions. These models might not be effective in preventing collisions when agents are nearby, especially if unexpected interactions occur due to environmental or mechanical factors. Several models were introduced for Safe Learning \cite{astghik2023_1}, \cite{astghik2023second}, using distributional robust control methods and safety layers that are computationally intensive. Implementing these methods on lightweight, real-time drone platforms can be challenging because of limited processing power and memory on board. 

The novelty of our research (shown in Fig. \ref{fig:title}) lies in addressing the intricate challenge of safe swarm landing in environments with dense crowd obstacles. The requirements of safe collision avoidance with obstacles significantly raise the difficulty of the landing task, even if we approximate the dynamics of the crowd as negligible and consider all obstacles as static. Our work focuses on real-time adaptability, where drones must navigate, avoid obstacles, and synchronize their landings on moving platforms without collision.
 
This adaptability is achieved by implementing collision avoidance, safe landing trajectory predictions, and continuous adjustment to moving targets, which are essential for real-world applications in cluttered settings. Extensive research has been conducted on drone path planning with MPC or Vision-based methods (e.g. April tags) \cite{krinner2024}, \cite{alireza2020}. However, existing methods may perform well under controlled or partially dynamic conditions and may not perform well when the dynamics of agents and obstacles are unknown. Our model is designed for environments where unpredictability and complexity are the norm, ensuring that drones can land safely and efficiently. \\

\section{System Overview}

\subsection{Motivation}
In an environment with obstacles, it is important to consider the safety of our system. We conducted experiments on drones with 2 to 4 obstacles with PPO and TD3 algorithms, also we used these algorithms with the Control Barrier Net function(CBF).
\subsection{Simulation Environment}
We used the gym-pybullet-drones simulation using our environment class, which includes a swarm of CrazyFlie drones, obstacles, and landing pads, which are small cylinders and cubes.
For training we used Core I9 computer with 64GB CPU with Nvidea RTX GeForce 4090 GPU with 24 GB VRAM. In the real environment, we used Vicon System which provides very high tracking accuracy for high-speed drones. 


\subsection{Designing the Reward Function}
Our designed reward is defined as follows:

\begin{equation}
r^i = r^i_{encourage} + r^i_{penalty} + r^i_{edge} + r^i_{velocity},
\label{eq:reward_equation}
\end{equation}
where $r^i_{encourage}$ is the reward for the i-th drone to come closer to its landing position, $r_{penalty}^i$ is the constant collision penalty for collision with the obstacles, $r_{edge}^i$ is the constant penalty for the drone being under its landing positions, $r^i_{velocity}$ is the penalty that encourages drones to decrease their velocity while being near to obstacles and the target. The reward $r^i_{encourage}$ can be calculated as follows:

\begin{equation}
r^i_{encourage} = \frac{\lambda}{\|P_{pad^i_j}\|_2^2 + \epsilon},
\label{eq:encourage_equation}
\end{equation}
where $\epsilon$, $\lambda$ are the constant coefficients, $P_{pad^i_j}$ are the coordinates of the i-th drone relative to the j-th target position. The $r^i_{velocity}$ can be calculated as follows:

\begin{equation}
r^i_{velocity} = \gamma \cdot max(v^i_{normed} \cdot (\alpha^i + \beta^i)),
\label{eq:vel_equation}
\end{equation}
where $\gamma$ is the constant that can be adjusted depending on the safety requirements, $v^i_{normed}$ is the normed velocity of i-th drone, $\alpha^i$ is the coefficient of the nearest obstacle, $\beta^i$ is the coefficient of the nearest target, which are defined as:

\begin{equation}
v^i_{normed} = \|v_i\|_2^2,
\end{equation}
\begin{equation}
\alpha^i = \max\left(\frac{1}{\|P_{\text{obs}_j^i}\|_2^2 + \epsilon}\right),
\end{equation}
\begin{equation}
\beta^i = \frac{1}{\|P_{pad_j^i}\|_2^2 + \epsilon},
\label{eq:vel_equation}
\end{equation}
where $\epsilon$ is the coefficient for stability of reward, $P_{\text{obs}_j^i}$ are the coordinates of the i-th drone relative to the j-th obstacle, $P_{pad_j^i}$ are the coordinates of the i-th drone relative to the j-th pad position.

\section{Experimental Evaluation}
The experiments were conducted using the same hardware setup as in the training phase.
\subsection{Simulation Results}
\begin{figure}[t]
\centering
\includegraphics[width=1.0\linewidth]{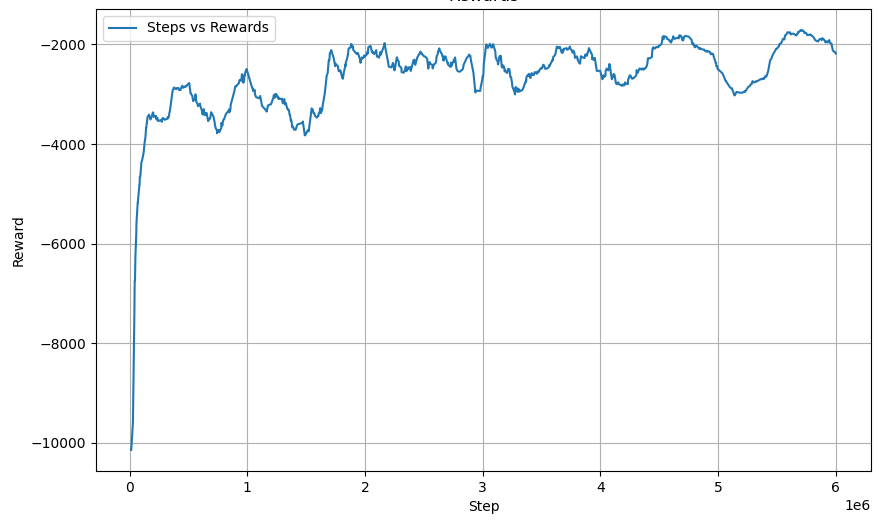}
\caption{Agent rewards during the training.}
\label{reward}
\end{figure}

We ran the landing task in simulation 30 times in total. The simulation revealed that SafeSwarm achieved 95\% accuracy of landing. The mean reward was -2000, as shown in Fig. \ref{reward}. 

 
\begin{figure}[t]
\centering
\includegraphics[width=1.0\linewidth]{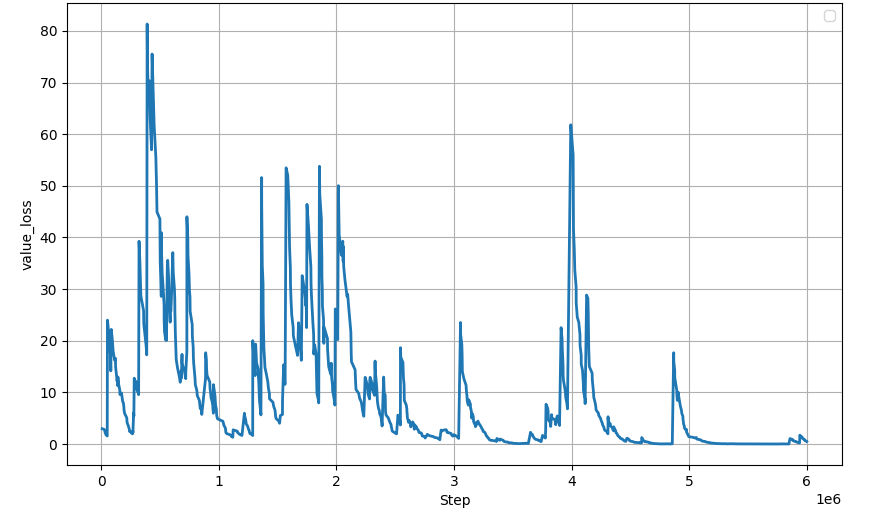}
\caption{Value loss during the training.}
\label{loss}
\end{figure}

The loss of value function reached 0.01 at the and of training, as shown in Fig. \ref{loss}. 

We have replicated the landing approach presented by \cite{aschu2024} and compared it with SafeSwarm in a simulated environment. Two conditions were tested: static and moving landing platforms with 3 obstacles around them. The results of the comparison under static conditions are presented in Table \ref{table:landing_precision}.

\begin{table}[H]
\centering
\caption{Comparison of SafeSwarm with MARLander, static platform}
\label{table:landing_precision}
\begin{tabular}{|l|c|c|c|} 
 \hline
 Algorithm & Success rate (\%) & Precision (cm) & Time (s) \\ 
 \hline
 MARLander & 91.67 & 2.26 & 12 \\ 
 \hline
 SafeSwarm & 95 & 2.25 & 17 \\ 
 \hline
\end{tabular}
\end{table}

The results of the comparison under dynamic conditions are presented in Table \ref{table:landing_precision2}.

\begin{table}[H]
\centering
\caption{Comparison of SafeSwarm with MARLander, moving platform}
\label{table:landing_precision2}
\begin{tabular}{|l|c|c|c|} 
 \hline
 Algorithm & Success rate (\%) & Precision (cm) & Time (s) \\ 
 \hline
 MARLander & 75 & 3.93 & 17 \\ 
 \hline
 SafeSwarm & 80 & 3.04 & 21 \\ 
 \hline
\end{tabular}
\end{table}

Overall, SafeSwarm showed higher precision in both static and dynamic conditions of the landing platform, however, the performance time was higher by 23\%.

\subsection{Real-World Experiment}
\begin{figure}[t]
\centering
\includegraphics[width=1\linewidth]{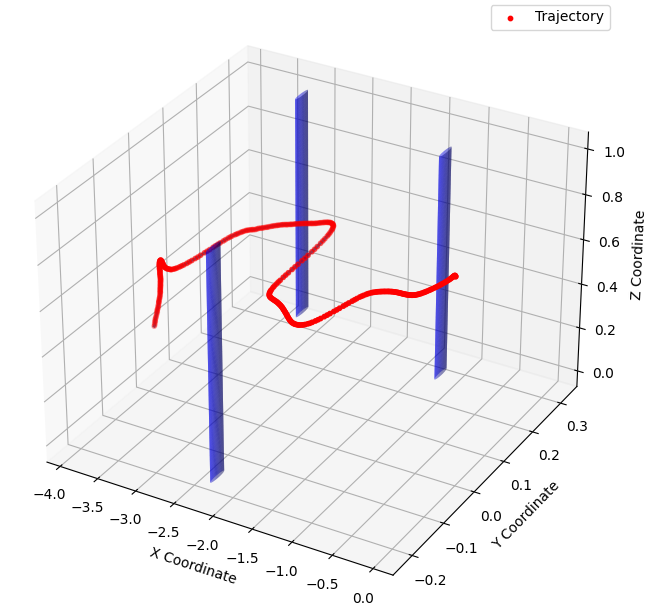}
\caption{3D trajectory for landing of our drone during real experiment.}
\label{fig: 3D trajectory}
\end{figure}

The experiment with Crazyflie 2.1 mini-quadrotor was performed in an indoor environment. The drone was initially located a 3 meter distance from the target with three obstacles on its path. Five test flights were conducted in total. The experimental results revealed the mean landing error of 2.25 cm and the accuracy of landing 80\% with a mean landing time of 17 s. The trajectory of the drone during one of the experiments is shown in Fig. \ref{fig: 3D trajectory}.

\section{Conclusion}

In this study, we focus on developing a Safe Reinforcement Learning framework for a swarm of drones to achieve a collision-free landing in environments with static obstacles. 
We achieved a landing accuracy of 2.25 cm with a mean time of 17 s in the real-world environment. 

Although our approach demonstrates promising results, our future work will include the incorporation of moving obstacles, the handling of complex obstacle patterns, and the addition of attention models to avoid collisions by incorporating attention mechanisms into the framework.
By addressing these aspects, we aim to enhance the safety, robustness, and versatility of swarm drone operations in increasingly complex and dynamic environments.

\balance
\bibliographystyle{IEEEtran}
\bibliography{references} 
\end{document}